\documentclass[10pt,twocolumn,letterpaper]{article}

\usepackage{cvpr}

\usepackage{graphicx}
\usepackage{amsmath}
\usepackage{amssymb}
\usepackage{booktabs}
\usepackage{graphicx}
\usepackage{multirow}
\usepackage{array, hhline}
\usepackage[table]{xcolor}

\usepackage[pagebackref,breaklinks,colorlinks]{hyperref}
\newcommand\blfootnote[1]{%
  \begingroup
  \renewcommand\thefootnote{}\footnote{#1}%
  \addtocounter{footnote}{-1}%
  \endgroup
}
\newcommand{\PAR}[1]{\vskip4pt \noindent{\bf #1~}}
\newcommand{\urlNewWindow}[1]{\href[pdfnewwindow=true]{#1}{\nolinkurl{#1}}}

\usepackage[capitalize]{cleveref}
\crefname{section}{Sec.}{Secs.}
\Crefname{section}{Section}{Sections}
\Crefname{table}{Table}{Tables}
\crefname{table}{Tab.}{Tabs.}

\def\cvprPaperID{60}
\def\confName{CVPR}
\def\confYear{2022}

\begin{document}

\title{Neural 3D Scene Reconstruction with the Manhattan-world Assumption}

\author{
    Haoyu Guo$^{1*}$
    \quad Sida Peng$^{1*}$
    \quad Haotong Lin$^{1}$
    \quad Qianqian Wang$^{2}$
    \\
    \quad Guofeng Zhang$^{1}$
    \quad Hujun Bao$^{1}$
    \quad Xiaowei Zhou$^{1\dagger}$
    \\
    $^1$ Zhejiang University \quad
    $^2$ Cornell University \quad
}

\maketitle

\blfootnote{The authors from Zhejiang University are affiliated with the State Key Lab of CAD\&CG and the ZJU-SenseTime Joint Lab of 3D Vision. $^*$Equal contribution. $^\dagger$Corresponding author: Xiaowei Zhou.}

\begin{abstract}
    This paper addresses the challenge of reconstructing 3D indoor scenes from multi-view images. Many previous works have shown impressive reconstruction results on textured objects, but they still have difficulty in handling low-textured planar regions, which are common in indoor scenes. An approach to solving this issue is to incorporate planer constraints into the depth map estimation in multi-view stereo-based methods, but the per-view plane estimation and depth optimization lack both efficiency and multi-view consistency. In this work, we show that the planar constraints can be conveniently integrated into the recent implicit neural representation-based reconstruction methods.
    Specifically, we use an MLP network to represent the signed distance function as the scene geometry. Based on the Manhattan-world assumption, planar constraints are employed to regularize the geometry in floor and wall regions predicted by a 2D semantic segmentation network. To resolve the inaccurate segmentation, we encode the semantics of 3D points with another MLP and design a novel loss that jointly optimizes the scene geometry and semantics in 3D space. Experiments on ScanNet and 7-Scenes datasets show that the proposed method outperforms previous methods by a large margin on 3D reconstruction quality. The code and supplementary materials are available at \urlNewWindow{https://zju3dv.github.io/manhattan\_sdf}.
\end{abstract}

\vspace{-0.2cm}
\section{Introduction}

Reconstructing 3D scenes from multi-view images is a cornerstone of many applications such as augmented reality, robotics, and autonomous driving. Given input images, traditional methods \cite{zheng2014patchmatch, schonberger2016pixelwise, schonberger2016structure} generally estimate the depth map for each image based on the multi-view stereo (MVS) algorithms and then fuse estimated depth maps into 3D models. Although these methods achieve successful reconstruction in most cases, they have difficulty in handling low-textured regions, e.g., floors and walls of indoor scenes, due to the unreliable stereo matching in these regions.

To improve the reconstruction of low-textured regions, a typical approach is leveraging the planar prior of man-made scenes, which has long been explored in literature \cite{coughlan1999manhattan,furukawa2009manhattan, romanoni2019tapa,gallup2010piecewise,xu2020planar,sun2021phi}.  
A renowned example is the Manhattan-world assumption \cite{coughlan1999manhattan}, i.e., the surfaces of man-made scenes should be aligned with three dominant directions.
These works either use plane estimation as a postprocessing step to inpaint the missing depth values in low-textured regions, or integrate planar constraints in stereo matching or depth optimization. However, all of them focus on optimizing per-view depth maps instead of the full scene models in 3D space. As a result, depth estimation and plane segmentation could still be inconsistent among views, yielding suboptimal reconstruction quality as demonstrated by our experimental results in Section \ref{section:result}. 

\begin{figure}[t]
\centering
\includegraphics[width=\linewidth]{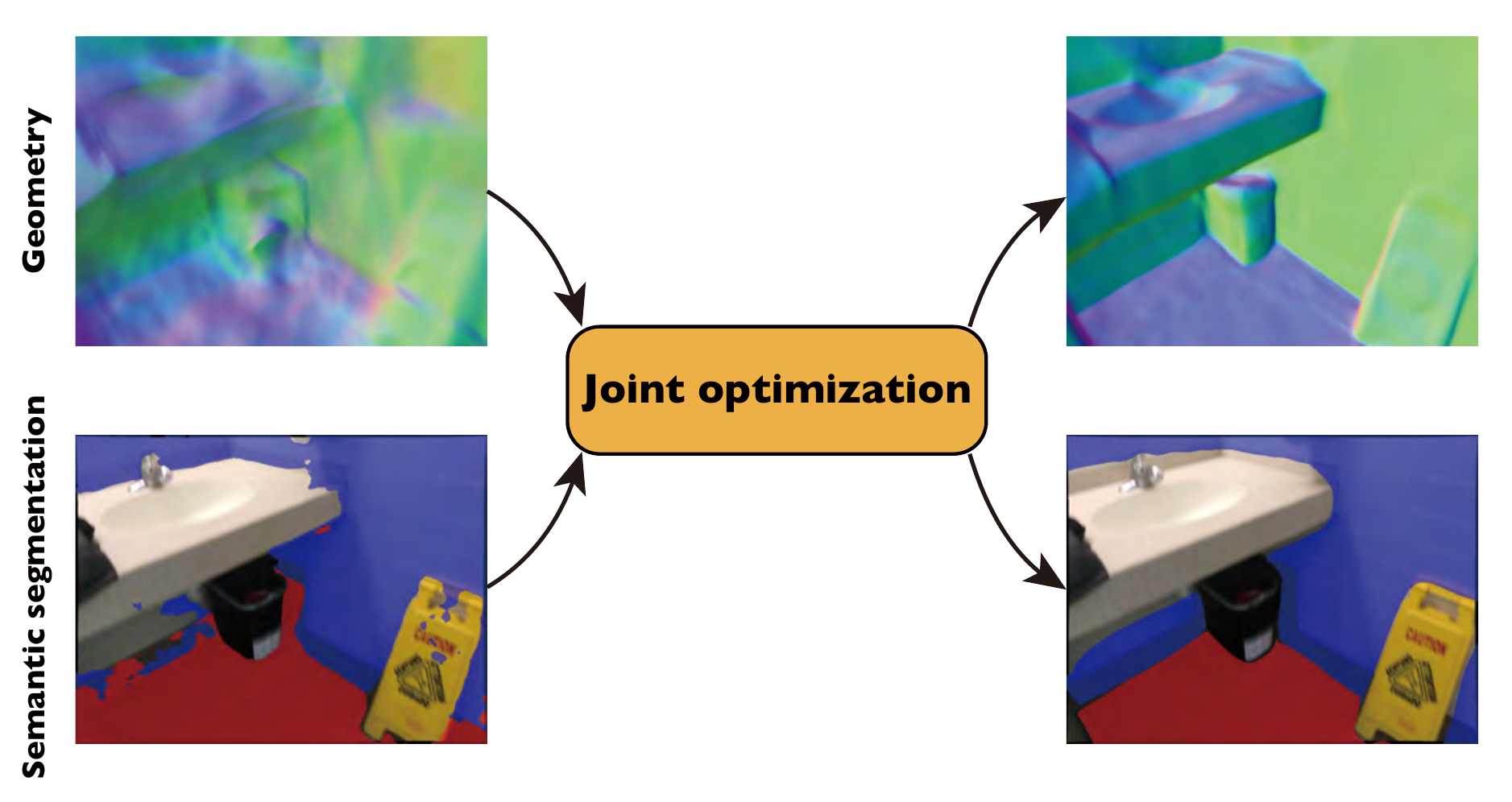}
\caption{\textbf{Core idea.} We represent the geometry and semantics of 3D scenes with implicit neural representations, which enables the joint optimization of geometry reconstruction and semantic segmentation in 3D space based on the Manhattan-world assumption.}
\label{fig:teaser}
\vspace{-0.2cm}
\end{figure}

There is a recent trend to represent 3D scenes as implicit neural representations \cite{sitzmann2019scene, niemeyer2020differentiable, yariv2020multiview} and learn the representations from images with differentiable renderers. In particular, \cite{yariv2020multiview, yariv2021volume, wang2021neus} use a signed distance field (SDF) to represent the scene and render it into images based on the sphere tracing or volume rendering. Thanks to the well-defined surfaces of SDFs, they recover high-quality 3D geometries from images. However, these methods essentially rely on the multi-view photometric consistency to learn the SDFs. So they still suffer from poor performance in low-textured planar regions, as shown in Figure~\ref{fig:teaser}, as many plausible solutions may satisfy the photometric constraint in low-textured planar regions.

In this work, we show that the Manhattan-world assumption \cite{coughlan1999manhattan} can be conveniently integrated into the learning of implicit neural representations of 3D indoor scenes and significantly improves the reconstruction quality. Unlike previous MVS methods that perform per-view depth optimization, implicit neural representations allow the joint representation and optimization of scene geometry and semantics simultaneously in 3D space, yielding globally-consistent reconstruction and segmentation.
Specifically, we use an MLP network to predict signed distance, color and semantic logits for any point in 3D space. The semantic logits indicate the probability of a point being floor, wall or background, initialized by a 2D semantic segmentation network \cite{chen2017deeplab}.
Similar to~\cite{yariv2021volume}, we learn the signed distance and color fields by comparing rendered images to input images based on volume rendering. 
For the surface points on floors and walls, we enforce their surface normals to respect the Manhattan-world assumption.
Considering the initial segmentation could be inaccurate, we design a loss that simultaneously optimizes the semantic logits along with the SDF. This loss effectively improves both the scene reconstruction and semantic segmentation, as illustrated in Figure~\ref{fig:teaser}. 

We evaluate our method on the ScanNet~\cite{dai2017scannet} and 7-Scenes~\cite{shotton2013scene} datasets, which are widely-used datasets for 3D indoor scene reconstruction.
The experiments show that the proposed approach outperforms 
the state-of-the-art methods in terms of reconstruction quality by a large margin, especially in planar regions.
Furthermore, the joint optimization of semantics and reconstruction improves the initial semantic segmentation accuracy.

In summary, our contributions are as follows:
\begin{itemize}
    \item A novel scene reconstruction approach that integrates the Manhattan-world constraint into the optimization of implicit neural representations.
    \item A novel loss function that optimizes semantic labels along with scene geometry.
    \item Significant gains of reconstruction quality compared to state-of-the-art methods on ScanNet and 7-Scenes.
\end{itemize}

\section{Related work}

\PAR{MVS.} Many methods adopt a two-stage pipeline for multi-view 3D reconstruction: first estimating the depth map for each image based on MVS and then performing depth fusion~\cite{merrell2007real,newcombe2011kinectfusion} to obtain the final reconstruction results.
Traditional MVS methods~\cite{schonberger2016structure,schonberger2016pixelwise} are able to reconstruct very accurate 3D shapes and have been used in many downstream applications such as novel view synthesis~\cite{Riegler2020FVS, Riegler2021SVS}.
However, they tend to give poor performance on texture-less regions.
A major reason is that texture-less regions make dense feature matching intractable.
To overcome this problem, some works improve the reconstruction pipeline with deep learning techniques.
For instance, 
~\cite{yao2018mvsnet,im2019dpsnet,yao2019recurrent} attempt to extract image features, build cost volumes and use 3D CNNs to predict depth maps.
~\cite{gu2020cascade,cheng2020deep} construct cost volumes in a coarse-to-fine manner and can achieve high resolution results.
Another line of works~\cite{xu2020planar,sun2021phi,romanoni2019tapa,gallup2010piecewise} utilize scene priors to help the reconstruction.
They observe that texture-less planar regions could be completed using planar prior.
\cite{kusupati2020normal,yin2019enforcing,long2020occlusion} propose a depth-normal consistency loss to improve training process.
Instead of predicting the depth map for each image, our method learns an implicit neural representation, which can achieve more coherent and accurate reconstruction. 

\PAR{Neural scene reconstruction.} Neural scene reconstruction methods predict the properties of points in the 3D space using neural networks.
Atlas~\cite{murez2020atlas} presents an end-to-end reconstruction pipeline which directly regresses truncated signed distance function from the 3D feature volume. 
NeuralRecon~\cite{sun2021neuralrecon} improves the reconstruction speed through reconstructing local surfaces for each fragment sequence.
They represent scenes as discrete voxels, resulting in the high memory consumption. 
Recently, some methods~\cite{mescheder2019occupancy,park2019deepsdf,sitzmann2019scene,mildenhall2020nerf,oechsle2021unisurf,wang2021neus,yariv2021volume, wei2021nerfingmvs} represent scenes with implicit neural functions and are able to produce high-resolution reconstruction with low memory consumption.
~\cite{niemeyer2020differentiable,liu2020dist} propose an implicit differentiable renderer, which enables learning 3D shapes from 2D images.
IDR~\cite{yariv2020multiview} models view-dependent appearance and can be applied to non-Lambertian surface reconstruction.
Despite achieving impressive performance, they need mask information to obtain the reconstruction.
Inspired by the success of NeRF~\cite{mildenhall2020nerf}, NeuS~\cite{wang2021neus} and VolSDF~\cite{yariv2021volume} attach volume rendering techniques to IDR and eliminate the need for mask information.
Although they achieve amazing reconstruction results of scenes with small scale and rich textures, we experimentally find that these methods tend to produce poor results in large scale indoor scenes with texture-less planar regions.
In contrast, our method utilizes semantic information to assist reconstruction in texture-less planar regions.

\begin{figure*}[t!]
\centering
\includegraphics[width=\linewidth]{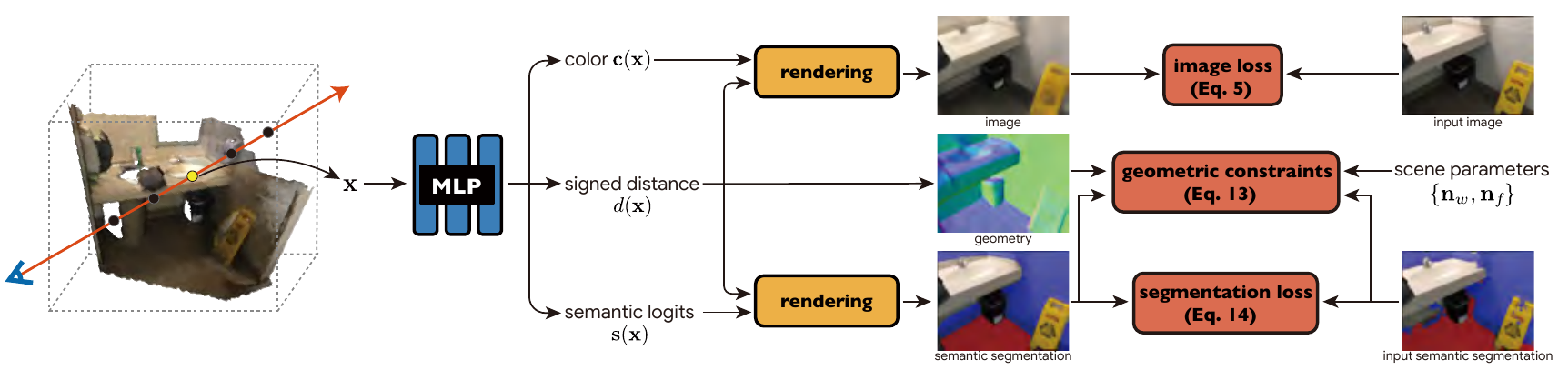}
\caption{\textbf{Overview of our method.} We learn the geometry, appearance and semantics of 3D scenes with implicit neural representations. For an image pixel, we use differentiable volume rendering to render its pixel color and semantic probabilities, which are supervised with input images and semantic labels in 2D. To jointly optimize the geometry and semantics, we introduce geometric constraints in planar regions based on the Manhattan-world assumption, which improves both the reconstruction and segmentation accuracy.}
\label{fig:pipeline}
\vspace{-0.3cm}
\end{figure*}

\PAR{Semantic segmentation.} Recently, learning-based methods have achieved impressive progress on semantic segmentation.
FCN~\cite{long2015fully} applies fully convolution on the whole image to produce pixel-level image semantic segmentation results.
Recent methods~\cite{badrinarayanan2017segnet,cheng2019spgnet} attempt to aggregate high-resolution feature maps using a learnable decoder to keep the detailed spatial information in the deep layers.
Another line of works~\cite{chen2017deeplab,zhao2017pyramid,chen2018encoder} use dilated convolutions for large receptive fields.
In addition to 2D segmentation methods, a lot of works aim to achieve semantic segmentation from 3D space.
~\cite{boulch2017unstructured, qi2017pointnet, qi2017pointnet++, qi2016volumetric} develop networks to process different representations of 3D data including point clouds and voxels.
More recently, ~\cite{zhi2021place} proposes to extend NeRF to encode semantics with radiance fields. The intrinsic multi-view consistency and smoothness of NeRF benefit semantics, which enables label propogation, super-resolution, denoising and several tasks. 
There are also some works~\cite{lawin2017deep, hu2021bidirectional,kundu2020virtual,liu20213d} that learn semantic segmentation in both 2D and 3D space and utilize the projection relation between images and 3D scenes to facilitate the performance.
Our method learns 3D semantics from 2D segmentation prediction~\cite{chen2018encoder} and jointly optimizes semantics with geometry.

\section{Method}

Given multi-view images with camera poses of an indoor scene, our goal is to reconstruct the high-quality scene geometry.
In this paper, we propose a novel approach called ManhattanSDF, as illustrated in Figure~\ref{fig:pipeline}.
We represent the scene geometry and appearance with signed distance and color fields, which are learned from images with volume rendering techniques (Sec.~\ref{sec:method_base}). 
To improve the reconstruction quality in texture-less regions (e.g., walls and floors), we perform semantic segmentation to detect these regions and apply the geometric constraints based on the Manhattan-world assumption~\cite{coughlan1999manhattan} (Sec.~\ref{sec:method_semantic}).
To overcome the inaccuracy of semantic segmentation, we additionally encode the semantic information into the implicit scene representation and jointly optimize the semantics together with the geometry and appearance of the scene (Sec.~\ref{sec:method_joint_opimization}).

\subsection{Learning scene representations from images}
\label{sec:method_base}

In contrast to MVS methods~\cite{schonberger2016pixelwise, yao2018mvsnet}, we model the scene as an implicit neural representation and learn it from images with a differentiable renderer.
Inspired by ~\cite{yariv2020multiview, yariv2021volume, wang2021neus}, we represent the scene geometry and appearance with signed distance and color fields. 
Specifically, given a 3D point $\mathbf{x}$, the geometry model maps it to a signed distance $d(\mathbf{x})$, which is defined as:
\begin{equation}
    (d(\mathbf{x}), \mathbf{z}(\mathbf{x}))  = F_d(\mathbf{x}),
\end{equation}
where $F_d$ is implemented as an MLP network, and $\mathbf{z}(\mathbf{x})$ is the geometry feature as in \cite{yariv2020multiview}.
To approximate the radiance function, the appearance model takes the spatial point $\mathbf{x}$, the view direction $\mathbf{v}$, the normal $\mathbf{n}(\mathbf{x})$, and the geometry feature $\mathbf{z}(\mathbf{x})$ as inputs and outputs color $\mathbf{c}(\mathbf{x})$, which is defined as:
\begin{equation}
    \mathbf{c}(\mathbf{x}) = F_\mathbf{c}(\mathbf{x}, \mathbf{v}, \mathbf{n}(\mathbf{x}), \mathbf{z}(\mathbf{x})),
\end{equation}
where we obtain the normal $\mathbf{n}(\mathbf{x})$ by computing the gradient of the signed distance $d(\mathbf{x})$ at point $\mathbf{x}$ as in \cite{yariv2020multiview}.

Following \cite{yariv2021volume, wang2021neus}, we adopt volume rendering to learn the scene representation networks from images.
Specifically, to render an image pixel, we sample $N$ points $\{\mathbf{x}_i\}$ along its camera ray $\mathbf{r}$. 
Then we predict the signed distance and color for each point.
To apply volume rendering techniques, we transform the signed distance $d(\mathbf{x})$ to the volume density $\sigma (\mathbf{x})$:
\begin{equation}
    \sigma(\mathbf{x})= \begin{cases} 
        \frac{1}{\beta}\left(1-\frac{1}{2}\exp{\left( \frac{d(\mathbf{x})}{\beta} \right)} \right) & \text{if } d(\mathbf{x})<0, \\
        \frac{1}{2 \beta} \exp{\left(-\frac{d(\mathbf{x})}{\beta} \right)} & \text{if } d(\mathbf{x})\geq 0,
\end{cases}
\end{equation}
where $\beta$ is a learnable parameter. Then we accumulate the densities and colors using numerical quadrature~\cite{mildenhall2020nerf}:
\begin{equation}
    \hat{{\mathbf{C}}}(\mathbf{r}) = \sum_{i=1}^{K}T_i(1-\exp(-\sigma_i \delta_i))\mathbf{c}_i,
\label{color_rendering}
\end{equation}
where $\delta_i = ||\mathbf{x}_{i + 1} - \mathbf{x}_i ||_2$ is the distance between adjacent sampled points, and $T_i=\exp(-\sum_{j=1}^{i-1}\sigma_j \delta_j)$ denotes the accumulated transmittance along the ray.

During training, we optimize the scene representation networks using multi-view images with photometric loss:
\begin{equation}
    \mathcal{L}_{\text{img}}= \sum_{\mathbf{r} \in \mathcal{R}} \left\lVert \hat{\mathbf{C}}(\mathbf{r}) - \mathbf{C}(\mathbf{r}) \right\rVert,
\label{eq:image}
\end{equation}
where $\mathbf{C}(\mathbf{r})$ is the ground-truth pixel color, and $\mathcal{R}$ is the set of camera rays going through sampled pixels.
Additionally, we apply Eikonal loss~\cite{gropp2020implicit} as suggested by~\cite{yariv2020multiview,yariv2021volume}.
\begin{equation}
    \mathcal{L}_{E}=\sum_{\mathbf{y} \in \mathcal{Y}}(\left\lVert \nabla_\mathbf{y} d(\mathbf{y}) \right\rVert_2 - 1)^2,
\label{eq:eikonal}
\end{equation}
where $\mathcal{Y}$ denotes the combination of points sampled from random uniform space and surface points for pixels.

\begin{figure}[t]
\centering
    \includegraphics[width=\linewidth]{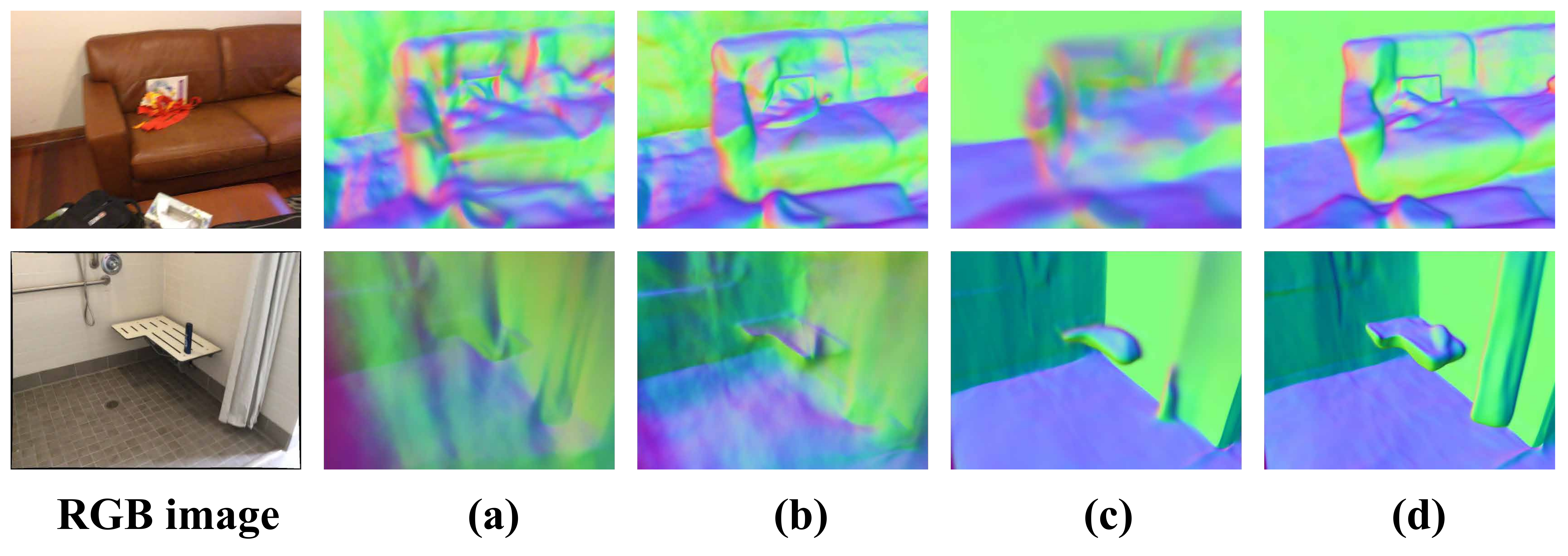}
    \caption{\textbf{Qualitative ablations.} (a) Training with only images. (b) Adding $\mathcal{L}_{d}$. (c) Adding $\mathcal{L}_{\text{geo}}$. (d) Replacing $\mathcal{L}_{\text{geo}}$ with $\mathcal{L}_{\text{joint}}$.}
\label{fig:normal}
\vspace{-0.3cm}
\end{figure}

We observe that learning the scene representation from scratch with only images has difficulty in reconstructing reasonable geometries even in textured regions, as shown in Figure~\ref{fig:normal}(a).
In contrast, although depth estimation based methods \cite{zheng2014patchmatch, schonberger2016pixelwise, schonberger2016structure} tend to give incomplete reconstructions in low-textured regions, they can reconstruct accurate point clouds of textured regions from images.
We propose to use depth maps from multi-view stereo method~\cite{schonberger2016structure} to assist the learning of the scene representations:
\begin{equation}
    \mathcal{L}_{d} = \sum_{\mathbf{r} \in \mathcal{D}} \left| \hat{D}(\mathbf{r}) - D(\mathbf{r}) \right|,
\end{equation}
where $\mathcal{D}$ is the set of camera rays going through image pixels that have depth values estimated by~\cite{schonberger2016structure}, $\hat{D}(\mathbf{r})$ and $D(\mathbf{r})$ are rendered and input depth values, respectively.  
Figure~\ref{fig:normal}(b) presents an example of the reconstruction result using the depth loss.
Although the depth loss improves the reconstruction quality, the reconstruction performance is still limited in texture-less regions, since input depth maps are incomplete in these regions. 

\subsection{Scene reconstruction with planar constraints}
\label{sec:method_semantic}

We observe that most texture-less planar regions lie on floors and walls.
As pointed by the Manhattan-world assumption~\cite{coughlan1999manhattan}, floors and walls of indoor scenes generally align with three dominant directions.
Motivated by this, we propose to apply the geometric constraints to the regions of floors and walls.
Specifically, we first use a 2D semantic segmentation network~\cite{chen2018encoder} to obtain the regions of floors and walls.
Then we apply loss functions to enforce the surface points in a planar region to share the same normal direction.

For the supervision of floor regions, we assume that floors are vertical to the z-axis following the Manhattan-world assumption.
We design the normal loss for a floor pixel as:
\begin{equation}
    \mathcal{L}_{f}(\mathbf{r}) = \left|1- \mathbf{n}(\mathbf{x}_\mathbf{r}) \cdot \mathbf{n}_{f}\right|,
\end{equation}
where $\mathbf{x}_\mathbf{r}$ is the surface intersection point of camera ray $\mathbf{r}$, $\mathbf{n}(\mathbf{x}_\mathbf{r})$ is the normal calculated as the gradient of signed distance $d(\mathbf{x})$ at point $\mathbf{x}_\mathbf{r}$, and $\mathbf{n}_{f}=\langle0,0,1\rangle$ is an upper unit vector that denotes the assumed normal direction in the floor regions. 

To supervise the wall regions, a learnable normal $\mathbf{n}_w$ is introduced.
We design a loss that enforces the normal directions of surface points on walls to be either parallel or orthogonal with the learnable normal $\mathbf{n}_w$, which is defined for wall pixels as:
\begin{equation}
    \mathcal{L}_{w}(\mathbf{r}) = \min_{i\in\{-1,0,1\}}\left|i- \mathbf{n}(\mathbf{x}_\mathbf{r}) \cdot \mathbf{n}_{w}\right|,
\end{equation}
where the learnable normal $\mathbf{n}_w$ is initialized as $\langle1,0,0\rangle$ and is jointly optimized with network parameters during training. We fix the last element of $\mathbf{n}_w$ as $0$ to force it vertical to $\mathbf{n}_{f}$.
Finally, we define the normal loss as:
\begin{equation}
    \mathcal{L}_{\text{geo}}=\sum_{r\in \mathcal{F}}\mathcal{L}_f(\mathbf{r})+\sum_{r\in \mathcal{W}}\mathcal{L}_w(\mathbf{r}),
\label{eq:normal_loss}
\end{equation}
where $\mathcal{F}$ and $\mathcal{W}$ are the sets of camera rays of image pixels that are predicted as floor and wall regions by the semantic segmentation network \cite{chen2018encoder}.

\subsection{Joint optimization of semantics and geometry}
\label{sec:method_joint_opimization}

Applying geometric constraints to floor and wall regions improves the reconstruction quality.
However, 2D semantic segmentation results predicted by the network could be wrong in some image regions, which leads to inaccurate reconstruction, as shown in Figure~\ref{fig:normal}(c).
To solve this problem, we propose to optimize semantic labels in 3D together with scene geometry and appearance.

Inspired by \cite{zhi2021place}, we augment the neural scene representation by additionally predicting semantic logits for each point in 3D space.
Let us denote semantic logits for $\mathbf{x}$ as $\mathbf{s}(\mathbf{x}) \in \mathbb{R}^3$.
The semantic logits are defined as:
\begin{equation}
    \mathbf{s}(\mathbf{x}) = F_\mathbf{s}(\mathbf{x}),
\end{equation}
where $F_\mathbf{s}$ is an MLP network.
By applying softmax function, the logits can be transformed to the probabilities of point $\mathbf{x}$ being floor, wall and other regions.
Similar to image rendering, we render the semantic logits into 2D image space with volume rendering techniques.
For an image pixel, its semantic logits are obtained by:
\begin{equation}
    \hat{{\mathbf{S}}}(\mathbf{r})=\sum_{i=1}^{N}T_i(1-\exp(-\sigma_i \delta_i))\mathbf{s}_i,
\end{equation}
where $\mathbf{s}_i$ is the logits of sampled point $\mathbf{x}_i$ along the camera ray $\mathbf{r}$.
We forward the logits $\hat{\mathbf{S}}$ into a softmax normalization layer to compute the multi-class probabilities $\hat{p}_f$, $\hat{p}_w$ and $\hat{p}_b$, denoting the probabilities of the pixel being floor, wall and other regions.

During training, we integrate the multi-class probabilities into the geometric losses proposed in Sec.~\ref{sec:method_semantic}.
To this end, we improve the normal loss in Equation~\eqref{eq:normal_loss} to a joint optimization loss, which is defined as:
\begin{equation}
    \mathcal{L}_\text{joint} = \sum_{\mathbf{r}\in \mathcal{F}} \hat{p}_f(\mathbf{r}) \mathcal{L}_f(\mathbf{r})  +
    \sum_{\mathbf{r}\in \mathcal{W}} \hat{p}_w(\mathbf{r}) \mathcal{L}_w(\mathbf{r}).
\end{equation}
 This loss function optimizes the scene representation in the following way.
Taking the floor region as an example,
if the input semantic label of $\mathbf{r}$ is correct, $\mathcal{L}_f(\mathbf{r})$ should decrease easily. But if the input segmentation is wrong, $\mathcal{L}_f(\mathbf{r})$ could vibrate during training. To decrease $\hat{p}_f(\mathbf{r}) \mathcal{L}_f(\mathbf{r})$, the gradient will push $\hat{p}_f(\mathbf{r})$ to be small, which thus optimizes the semantic label.
Note that a trivial solution is that both $\hat{p}_f$ and $\hat{p}_w$ vanish. To avoid this, we also supervise the semantics with input semantic segmentation results estimated by \cite{chen2018encoder} using the cross entropy loss:
\begin{equation}
    \mathcal{L}_\mathbf{s} = - \sum_{\mathbf{r} \in \mathcal{R}} \sum_{k \in \{f,w,b\}} p_{k}(\mathbf{r})\log \hat{p}_{k}(\mathbf{r}),
\end{equation}
where $\hat{p}_k(\mathbf{r})$ is the rendered probability for class $k$ and $p_k(\mathbf{r})$ is 2D semantic segmentation prediction.
Note that learning 3D semantics with $\mathcal{L}_\mathbf{s}$ naturally utilizes the multi-view consensus to improve the accuracy of semantic scene segmentation, as shown in \cite{zhi2021place}.

\section{Implementation details}

We implement our method with PyTorch~\cite{paszke2019pytorch} and adopt DeepLabV3+~\cite{chen2018encoder} from Detectron2~\cite{wu2019detectron2} for implementing 2D semantic segmentation network.
The network training is performed on one NVIDIA TITAN Xp GPU.
In experiments, we first normalize all cameras to be inside a unit sphere and initialize network parameters following~\cite{atzmon2020sal} so that the SDF is approximated to a uint sphere, and the surface normals of the sphere are facing inside.
Images are resized to $640\times480$ for both 2D semantic segmentation and scene reconstruction. We use Adam optimizer~\cite{kingma2014adam} with learning rate of 5e-4 and train the network with batches of 1024 rays for 50k iterations. The optimization can be completed in about 5 hours for each scene.
We use Marching Cubes algorithm~\cite{lorensen1987marching} for extracting surface mesh from the learned signed distance function.

\begin{figure*}[t!]
\centering
\includegraphics[width=\linewidth]{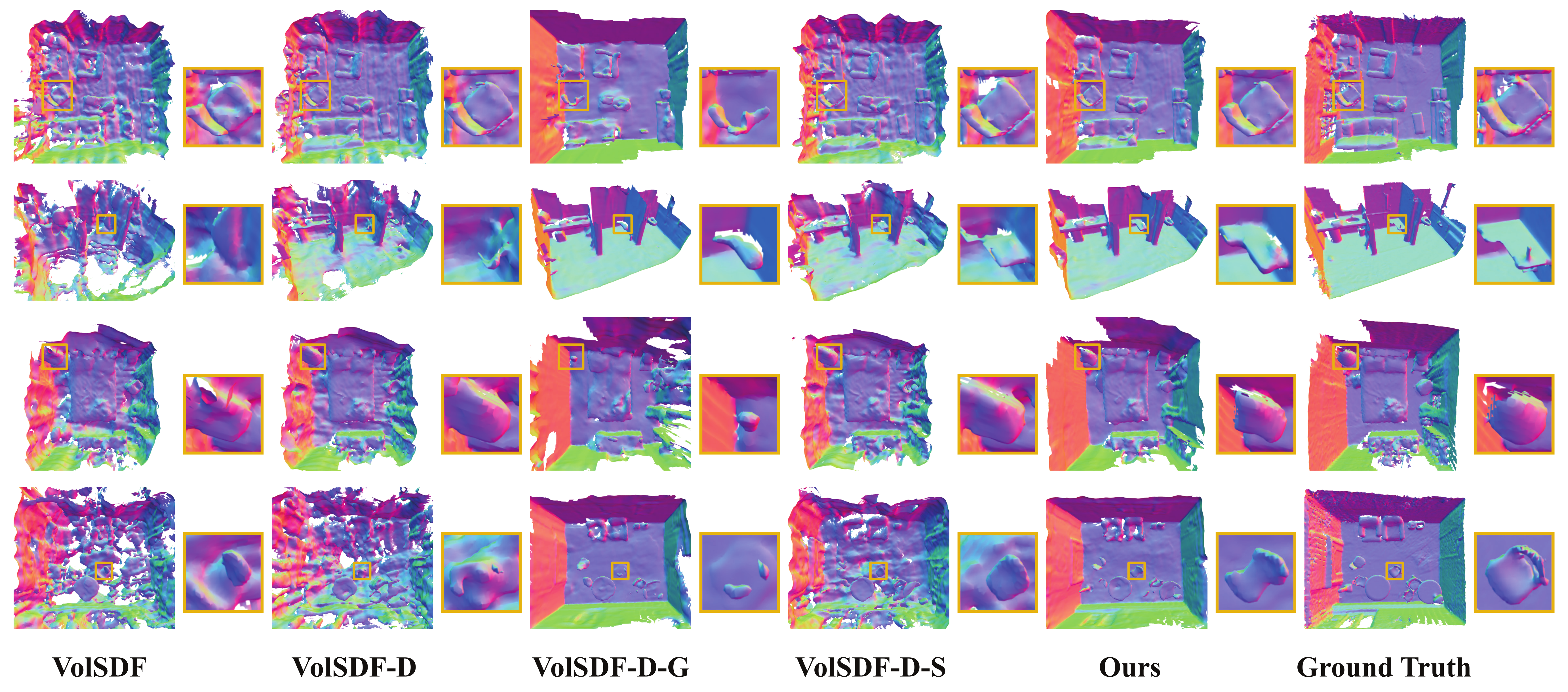}
\caption{\textbf{Ablation studies on ScanNet.} Our method can produce much more coherent reconstruction results compared to our baselines. Note that \textbf{VolSDF-D-G} can reconstruct smoother and more complete planes compared to \textbf{VolSDF} and \textbf{VolSDF-D}. \textbf{Ours} can maintain the reconstruction quality of planes while also reconstruct much more details in non-planar regions compared to \textbf{VolSDF-D-G}. The color indicates surface normal. Zoom in for details.}
\label{fig:ablation_scannet}
\vspace{-0.3cm}
\end{figure*}
\section{Experiments}

\subsection{Datasets, metrics and baselines}

\PAR{Datasets.} We perform the experiments on ScanNet (V2) \cite{dai2017scannet} and 7-Scenes \cite{shotton2013scene}. ScanNet is an RGB-D video dataset that contains 1613 indoor scenes with 2.5 million views. It is annotated with ground-truth camera poses, surface reconstructions, and instance-level semantic segmentations. 7-Scenes consists of RGB-D frames recorded by a handheld Kinect RGB-D camera. It uses KinectFusion to obtain camera poses and dense 3D models. In our experiments, we train the 2D semantic segmentation network on training set of ScanNet and perform the experiments on 8 randomly selected scenes (4 from validation set of ScanNet and 4 from 7-Scenes). Each scene contains $1\text{K} \text{--}5\text{K} $ views. We uniformly sample one tenth views for reconstruction.

\PAR{Metrics.}
For 3D reconstruction, we use RGB-D fusion results as ground truth and evaluate our method using 5 standard metrics defined in ~\cite{murez2020atlas}: accuracy, completeness, precision, recall and F-score. We consider F-score as the overall metric following~\cite{sun2021neuralrecon}.
The definitions of these metrics are detailed in the supplementary material.
For semantic segmentation, we evaluate Intersection over Union (IoU) of floor and wall.

\PAR{Baselines.} (1) Classical MVS method: COLMAP~\cite{schonberger2016structure}. We use screened Poisson Surface reconstruction (sPSR)~\cite{kazhdan2013screened} to reconstruct mesh from point clouds. (2) MVS methods with plane fitting: COLMAP*. There are several methods~\cite{romanoni2019tapa,gallup2010piecewise} that segment piece-wise plane segmentations in image space and apply plane fitting to COLMAP. Since these methods have not released code, we implement this baseline using state-of-the-art piece-wise plane segmentation method~\cite{liu2019planercnn} and denote it as COLMAP$^*$. (3) MVS method with plane regularization: ACMP~\cite{xu2020planar}. ACMP utilizes a probabilistic graphical model to embed planar models into PatchMatch and proposes multi-view aggregated matching cost to improve depth estimation in planar regions. (4) State-of-the-art volume rendering based methods: NeRF~\cite{mildenhall2020nerf}, UNISURF~\cite{oechsle2021unisurf}, NeuS~\cite{wang2021neus} and VolSDF~\cite{yariv2021volume}. For these methods, we use Marching Cubes algorithm~\cite{lorensen1987marching} to extract mesh. Since they (including our method) can reconstruct unobserved regions which will be penalized in evaluation, we render depth maps from predicted mesh and re-fuse them using TSDF fusion~\cite{newcombe2011kinectfusion} following ~\cite{murez2020atlas}.

\subsection{Ablation studies}
\label{section:ablation}

We conduct ablation studies on ScanNet and show the effectiveness of each component in our method.

We train with four configurations: (1) raw setting of \textbf{VolSDF}: training network with only image supervision, (2) \textbf{VolSDF-D}: we add depth supervision $\mathcal{L}_d$ defined in Section~\ref{sec:method_base}, (3) \textbf{VolSDF-D-G}: in addition to VolSDF-D, we add normal loss $\mathcal{L}_{\text{geo}}$ defined in Section~\ref{sec:method_semantic}, (4) \textbf{VolSDF-D-S}: in addition to VolSDF-D, we learn semantics in 3D space, (5) \textbf{Ours}: we learn semantics in 3D space and improve normal loss to joint optimization loss $\mathcal{L}_{\text{joint}}$ defined in Section~\ref{sec:method_joint_opimization}.
We report quantitative results in Table~\ref{tab:ablation} and provide qualitative results in Figure~\ref{fig:ablation_scannet}.

\begin{table}[t!]
\centering
\setlength{\tabcolsep}{1.5mm}{
\begin{tabular}{lccccc}
\toprule
& Acc$\downarrow$ & Comp$\downarrow$ & Prec$\uparrow$ & Recall$\uparrow$ & F-score$\uparrow$ \\
\midrule
VolSDF     & 0.414 & 0.120 & 0.321 & 0.394 & 0.346  \\
VolSDF-D   & 0.229 & 0.099 & 0.416 & 0.455 & 0.431  \\
VolSDF-D-G & 0.133 & 0.090 & 0.447 & 0.435 & 0.438  \\
VolSDF-D-S & 0.127 & 0.081 & 0.463 & 0.487 & 0.474  \\
Ours       & \textbf{0.072} & \textbf{0.068} & \textbf{0.621} & \textbf{0.586} & \textbf{0.602}  \\
\bottomrule
\end{tabular}
}
\caption{\textbf{Ablation studies on ScanNet.} We report 3D reconstruction metrics. Our method has a notable improvement in terms of both accuracy and completeness compared to our baselines.}
\label{tab:ablation}
\vspace{-0.2cm}
\end{table}

\begin{figure*}[t!]
\centering
    \includegraphics[width=\linewidth]{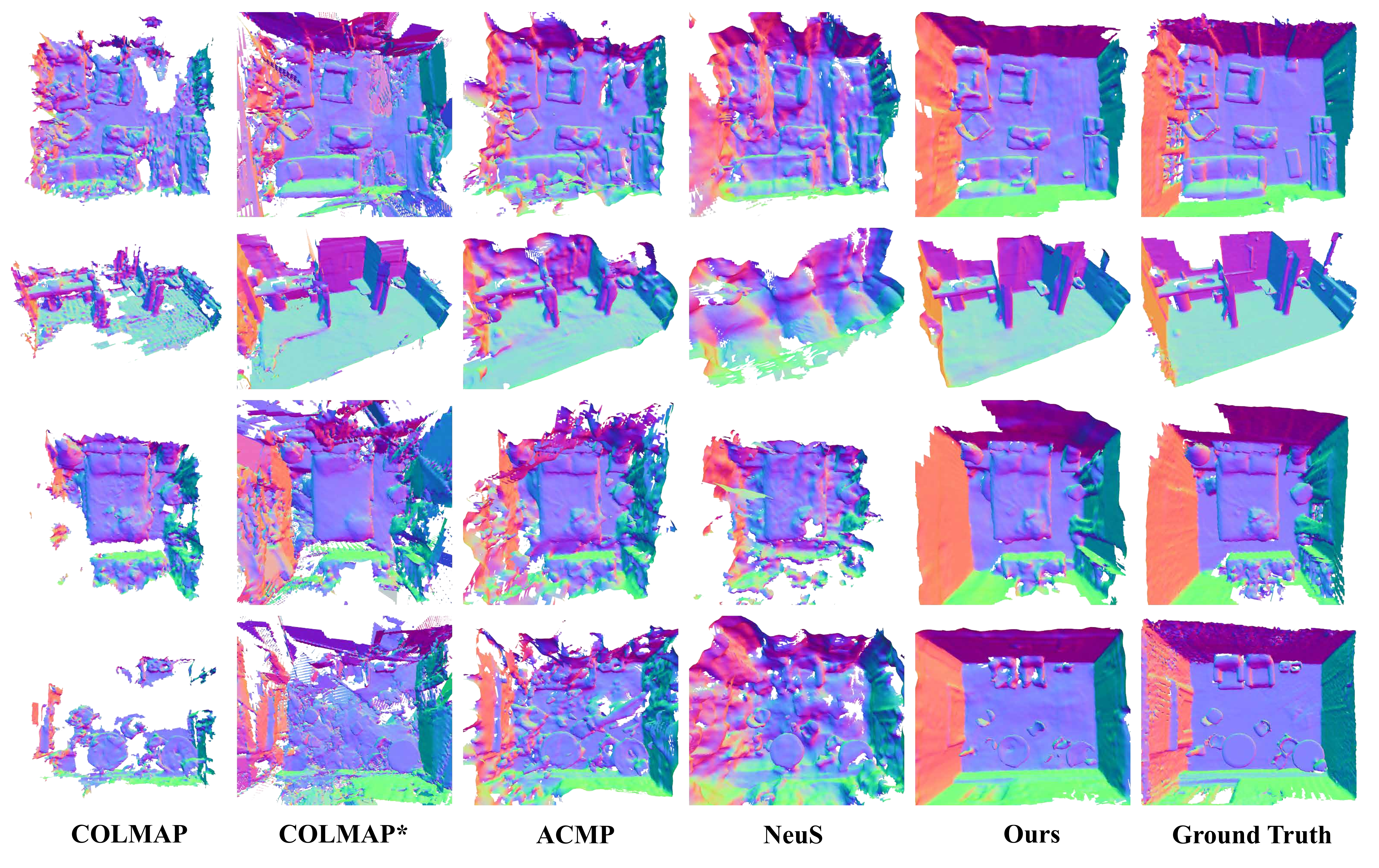}
    \caption{\textbf{3D reconstruction results on ScanNet.} Our method significantly outperforms COLMAP and volume rendering-based methods. Furthermore, compared with methods that apply planar prior to MVS, we can produce more coherent reconstruction results especially in planar regions. Zoom in for details.}
    \label{fig:qual_geo_scannet}
\end{figure*}

\begin{table*}[t!]
\centering
\scalebox{1.0}{
\begin{tabular}{l|ccccc|ccccc}
\hline

\multirow{2}{*}{Method} & \multicolumn{5}{c|}{ScanNet} & \multicolumn{5}{c}{7-Scenes} \\
\hhline{~|----------}
& \multicolumn{1}{c}{Acc$\downarrow$} & \multicolumn{1}{c}{Comp$\downarrow$} & \multicolumn{1}{c}{Prec$\uparrow$} & \multicolumn{1}{c}{Recall$\uparrow$} & \multicolumn{1}{c|}{\cellcolor{gray!25}\textbf{F-score}$\uparrow$} & \multicolumn{1}{c}{Acc$\downarrow$} & \multicolumn{1}{c}{Comp$\downarrow$} & \multicolumn{1}{c}{Prec$\uparrow$} & \multicolumn{1}{c}{Recall$\uparrow$} & \multicolumn{1}{c}{\cellcolor{gray!25}\textbf{F-score}$\uparrow$} \\

\hhline{-----------}

COLMAP     & \textbf{0.047} & 0.235 & \textbf{0.711} & 0.441 & \cellcolor{gray!25}0.537  & \textbf{0.069} & 0.417 & \textbf{0.536} & 0.202 & \cellcolor{gray!25}0.289 \\
COLMAP$^*$ & 0.396 & 0.081 & 0.271 & \textbf{0.595} & \cellcolor{gray!25}0.368  & 0.670 & 0.215 & 0.116 & 0.215 & \cellcolor{gray!25}0.149 \\
ACMP       & 0.118 & 0.081 & 0.531 & 0.581 & \cellcolor{gray!25}0.555  & 0.293 & 0.194 & 0.350 & 0.269 & \cellcolor{gray!25}0.299 \\
NeRF       & 0.735 & 0.177 & 0.131 & 0.290 & \cellcolor{gray!25}0.176  & 0.573 & 0.321 & 0.159 & 0.085 & \cellcolor{gray!25}0.083 \\
UNISURF    & 0.554 & 0.164 & 0.212 & 0.362 & \cellcolor{gray!25}0.267  & 0.407 & 0.136 & 0.195 & 0.301 & \cellcolor{gray!25}0.231 \\
NeuS       & 0.179 & 0.208 & 0.313 & 0.275 & \cellcolor{gray!25}0.291  & 0.151 & 0.247 & 0.313 & 0.229 & \cellcolor{gray!25}0.262 \\
VolSDF     & 0.414 & 0.120 & 0.321 & 0.394 & \cellcolor{gray!25}0.346  & 0.285 & 0.140 & 0.220 & 0.285 & \cellcolor{gray!25}0.246 \\ \hhline{-----------}
Ours       & 0.072 & \textbf{0.068} & 0.621 & 0.586 & \cellcolor{gray!25}\textbf{0.602}  & 0.112 & \textbf{0.133} & 0.351 & \textbf{0.326} & \cellcolor{gray!25}\textbf{0.336} \\
\hhline{-----------}

\end{tabular}
}
\caption{\textbf{Averaged 3D reconstruction metrics on ScanNet and 7-Scenes.} We compare our method with MVS and volume rendering based methods. The accuracy of our method ranks only second to COLMAP and our completeness is on par with MVS methods with planar prior. Considering both accuracy and completeness, our method achieves the best reconstruction performance.}
\label{tab:qual_geo}
\vspace{-0.2cm}
\end{table*}

Comparing \textbf{VolSDF} and \textbf{VolSDF-D} in Table~\ref{tab:ablation} shows that supervision from estimated sparse depth maps gives about 0.095 precision improvement and 0.061 recall improvement. Visualization results in Figure~\ref{fig:ablation_scannet} show that there are improvements in both planar and non-planar regions, but the reconstruction is still noisy and incomplete. These results demonstrate that $\mathcal{L}_d$ can make network converge better but the reconstruction results are still of low quality.

Then, we study how the normal loss affects the reconstruction performance. Results in Table~\ref{tab:ablation} show that \textbf{VolSDF-D-G} gives 0.031 precision improvement, but recall is decreased by 0.020. As shown in visualization results in Figure~\ref{fig:ablation_scannet}, \textbf{VolSDF-D-G} can reconstruct smoother and more complete planes compared to \textbf{VolSDF-D}, but some details of non-planar regions are missed. These results demonstrate that $\mathcal{L}_{\text{geo}}$ can improve the reconstruction in planar regions, but the performance in non-planar regions could be decreased due to misleading of wrong segmentation.

To validate the benefit of learning semantic fileds, we compare \textbf{VolSDF-D} and \textbf{VolSDF-D-S}. Results in Table~\ref{tab:ablation} show that \textbf{VolSDF-D-S} gives 0.047 precision improvement and 0.032 recall improvement. These results demonstrate that learning semantics in 3D space can also assist reconstruction.

To validate the benefit of our proposed joint optimization manner, we compare \textbf{VolSDF-D-G} and \textbf{Ours} in Table~\ref{fig:ablation_scannet}. Substituting $\mathcal{L}_{\text{geo}}$ with $\mathcal{L}_{\text{joint}}$ gives 0.174 precision improvement and 0.151 recall improvement. Visualization results in Figure~\ref{fig:ablation_scannet} show that, while \textbf{Ours} can keep great reconstruction performance in planar regions, the reconstruction in non-planar regions are also improved notably. These results demonstrate that \textbf{Ours} can achieve the most coherent reconstruction results.

\subsection{Comparisons with the state-of-the-art methods}
\label{section:result}

\PAR{3D reconstruction.} We evaluate 3D geometry metrics on ScanNet and 7-Scenes. Averaged quantitative results are shown in Table~\ref{tab:qual_geo}. Please refer to the supplementary material for detailed results on individual scenes.
Qualitative results on ScanNet are shown in Figure~\ref{fig:qual_geo_scannet}.
By analysing quantitative and qualitative results, we found that our method significantly outperforms state-of-the-art MVS and volume rendering based methods considering both reconstruction precision and recall.

COLMAP can achieve extremely high precision as it filters out reconstructed points which are inconsistent between multiple views in fusion stage. However, this process sacrifices recall.
COLMAP$^*$ and ACMP can obviously complete some missing areas and acheive higher recall by applying planar prior to COLMAP. However, their optimization strategy can not guarantee the consistency of estimated depth maps, resulting in noisy reconstructions.
The performance of NeRF is poor since the volume density representation has not sufficient constraint on geometry.
Other volume rendering based methods -- UNISURF, NeuS and VolSDF perform better than NeRF as occupancy and signed distance function have better surface constraints. However, they still struggle in reconstructing accurate and complete geometry.

\begin{figure}[t!]
\centering
    \includegraphics[width=\linewidth]{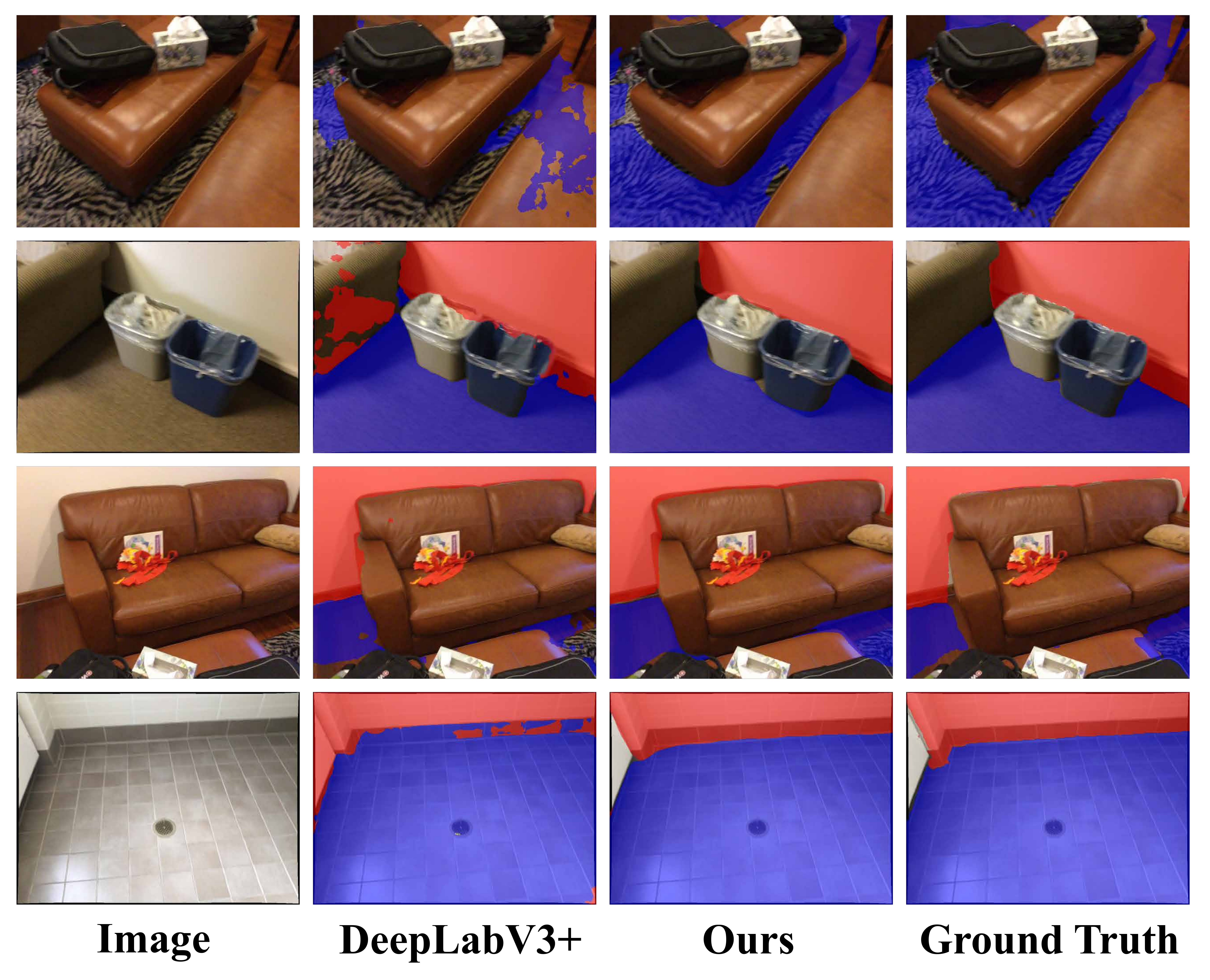}
    \caption{\textbf{Semantic segmentation results.} We compare our semantic segmentation results with DeepLabV3+. We mask pixels of floor and wall labels with blue and red.}
    \label{fig:semantic}
\end{figure}

\begin{table}[t]
\centering
\begin{tabular}{lccc}
\toprule
& IoU$^f\uparrow$ & IoU$^w\uparrow$ & IoU$^{m}\uparrow$  \\
\midrule
DeepLabV3+ & 0.532 & 0.475 & 0.503  \\
Ours       & \textbf{0.624} & \textbf{0.518} & \textbf{0.571}  \\  
\bottomrule
\end{tabular}
\caption{\textbf{Quantitative results of semantic segmentation.} IoU$^{f}$ and IoU$^{w}$ denote IoU of floor and wall regions, respectively. IoU$^{m}$ denotes the average of IoU$^{f}$ and IoU$^{w}$.}
\label{tab:semantic}
\vspace{-0.2cm}
\end{table}

\PAR{Semantic segmentation.} We render our learned semantics to image space and evaluate semantic segmentation metrics on ScanNet. We compare our method with DeepLabV3+. and report quantitative results in Table~\ref{tab:semantic}.
Qualitative results are shown in Figure~\ref{fig:semantic}.
Quantitatively, our metrics in both floor and wall regions are improved distinctly compared to DeepLabV3+.
Visualization results show that semantic segmentation results predicted by DeepLabV3+ have non-negligible noise especially near boundaries.
The noise are ruleless and  generally inconsistent between different views.
By learning semantics in 3D space, our method can naturally combine multi-view information and improve consistency so that the noise could be remitted notably.
However, there are also some misclassified pixels that cannot be easily corrected using multi-view consistency. Taking the last row of Figure~\ref{fig:semantic} for example, the bottom of wall has different color with the main part of the wall, so that some pixels are wrongly recognized as floor.
This kind of phenomenon can occur in every view and is different from the inconsistent noise.
By optimizing semantics together with geometry, these misclassified pixels could be corrected.

\PAR{Novel view synthesis.} Our accurate reconstruction results enable us to render high-quality images under novel views. To evaluate novel view synthesis, we select some novel views away from training views and render images. The qualitative comparison are shown in Figure~\ref{fig:view_synthesis}. More results can be found in the supplementary material.

\begin{figure}[t!]
\centering
    \includegraphics[width=\linewidth]{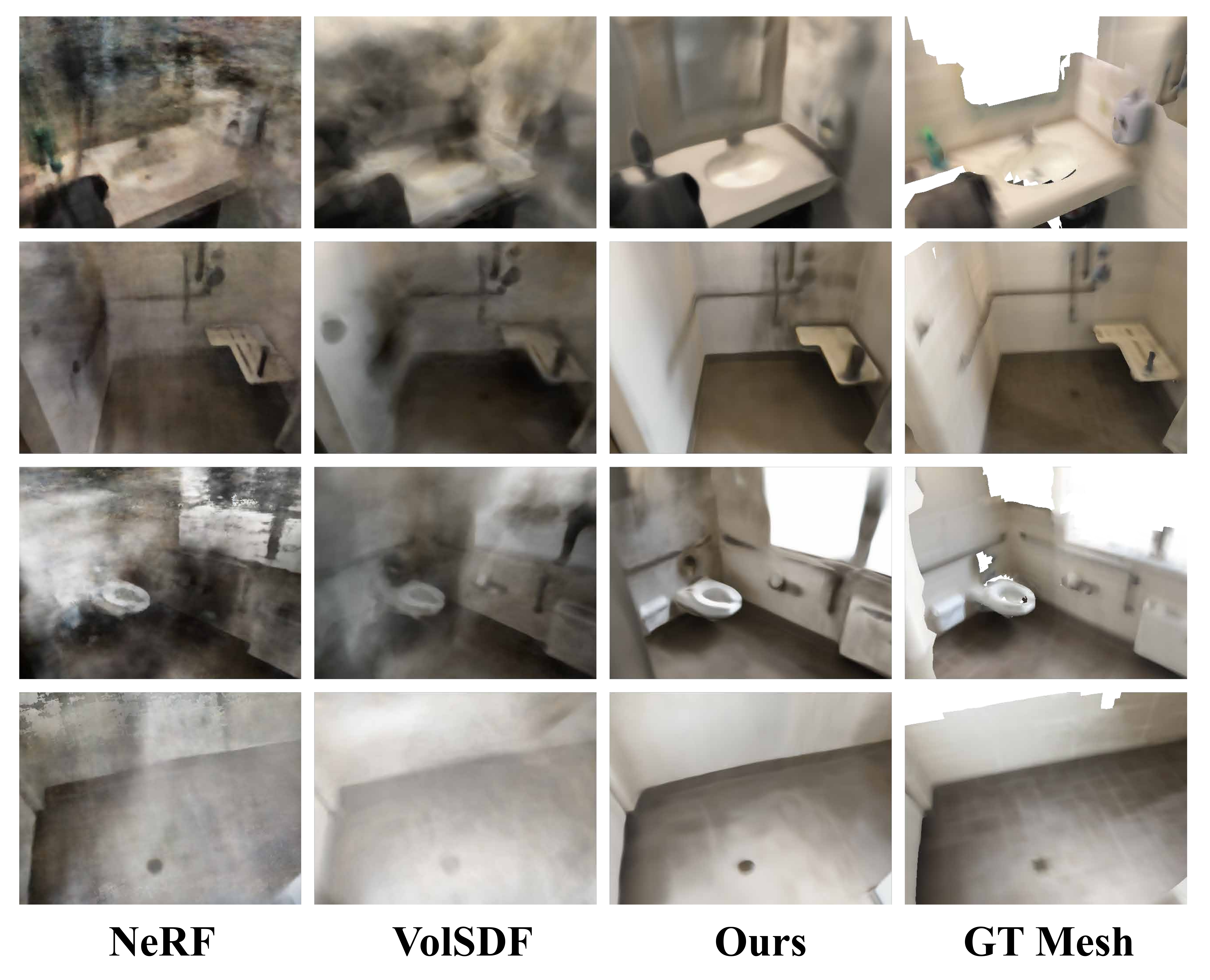}
    \caption{\textbf{Novel view synthesis results.} We select novel views relatively far from training views for the qualitative comparison. Our method produces better rendering results compared to NeRF and VolSDF. Due to the lack of ground truth images in novel views, we render GT mesh in these views for reference.}
    \label{fig:view_synthesis}
    \vspace{-0.3cm}
\end{figure}

\section{Conclusion}

In this paper, we introduced a novel indoor scene reconstruction method based on the Manhattan-world assumption.
The key idea is to utilize semantic information in planar regions to guide geometry reconstruction.
Our method learns 3D semantics from 2D segmentation results, and jointly optimizes 3D semantics with geometry to improve the robustness against inaccurate 2D segmentation.
Experiments showed that the proposed method was able to reconstruct accurate and complete planes while maintaining details of non-planar regions, and significantly outperformed the state-of-the-art methods on public datasets.

\PAR{Limitations.} This work only considers the Manhattan-world assumption.
While most man-made scenes obey this assumption, some scenarios require a more general assumption, e.g., the Atlanta-world assumption~\cite{schindler2004atlanta}. The proposed framework could be extended to adopt other assumptions by modifying the formulation of geometric constraints in the loss function.

\PAR{Acknowledgement.}
The authors would like to acknowledge the support from the National Key Research and Development Program of China (No. 2020AAA0108901), NSFC (No. 62172364), and the ZJU-SenseTime Joint Lab of 3D Vision.

{\small
\bibliographystyle{ieee_fullname}
\bibliography{egbib}
}

\newpage

\section*{Appendix}
\appendix

\section{Network architecture}

\begin{figure}[h]
\centering
\includegraphics[width=\linewidth]{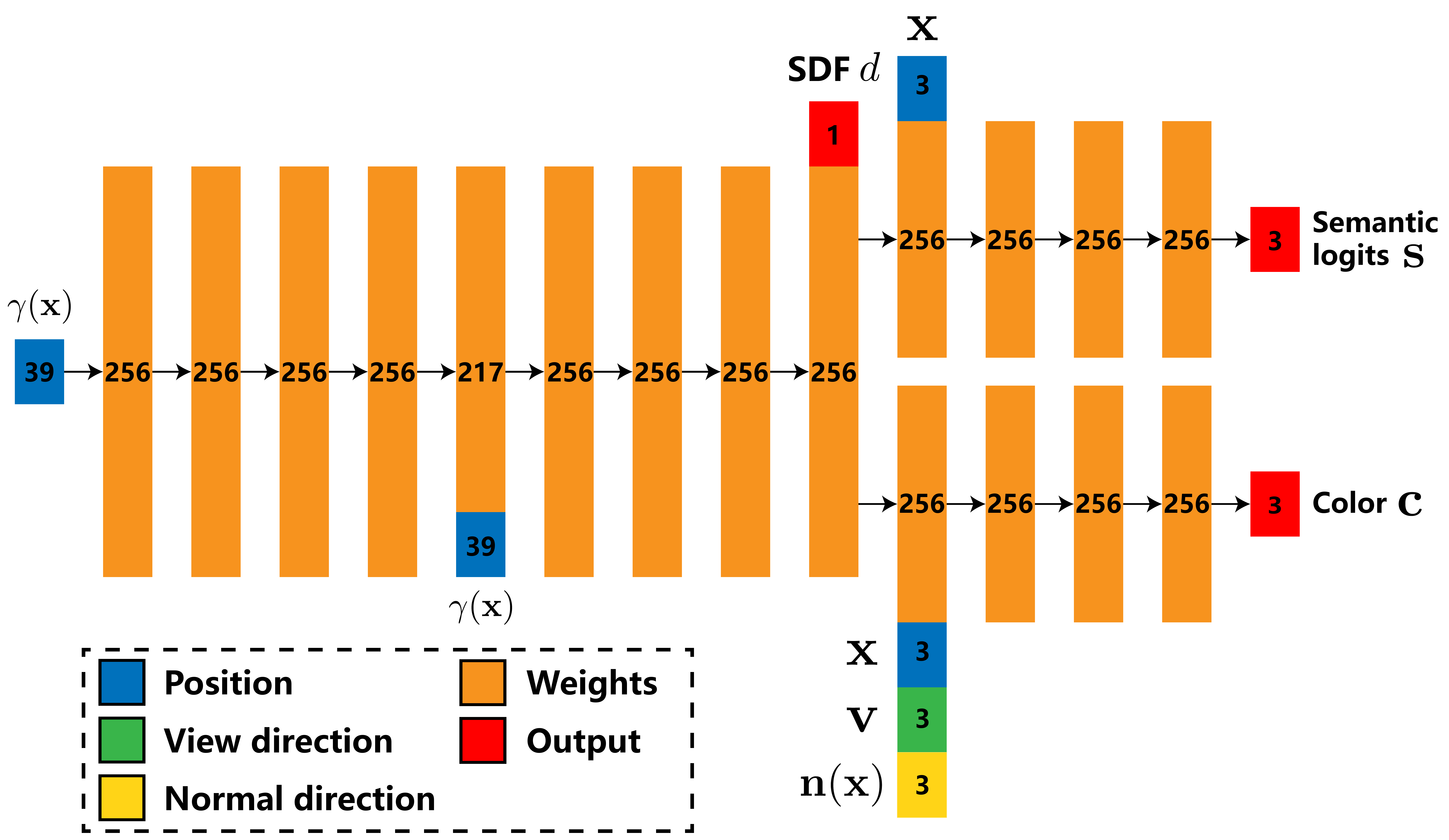}
\caption{\textbf{Network architecture.} Our network takes as inputs the spatial point $\mathbf{x}$, the view direction $\mathbf{v}$, and outputs SDF $d$, color $\mathbf{c}$, semantic logits $\mathbf{s}$.}
\label{fig:arch}
\vspace{-0.5cm}
\end{figure}

The architecture of our network is illustrated in Figure~\ref{fig:arch}.

\vspace{-0.2cm}

\section{Evaluation metrics}

The definitions of 3D reconstruction metrics are shown in Table~\ref{tab:metric_defs}.

\begin{table}[ht]
\centering
\begin{tabular}{lc}
\toprule
Metric & Definition \\
\midrule
Acc & $\mbox{mean}_{p \in P}(\min_{p^*\in P^*}||p-p^*||)$ \\
Comp & $\mbox{mean}_{p^* \in P^*}(\min_{p\in P}||p-p^*||)$ \\
Prec & $\mbox{mean}_{p \in P}(\min_{p^*\in P^*}||p-p^*||<.05)$ \\
Recal & $\mbox{mean}_{p^* \in P^*}(\min_{p\in P}||p-p^*||<.05)$ \\
F-score & $\frac{ 2 \times \text{Perc} \times \text{Recal} }{\text{Prec} + \text{Recal}}$ \\
\bottomrule
\end{tabular}
\caption{\textbf{Metric definitions.} $P$ and $P^*$ are the point clouds sampled from predicted and ground truth mesh.}
\vspace{-0.5cm}
\label{tab:metric_defs}
\end{table}

\section{Quantitative results on individual scenes}

We conduct reconstruction experiments on 8 randomly select scenes from ScanNet and 7-Scenes datasets, and compare our method with state-of-the-art MVS and implicit neural representations based methods. We show quantitative results on each individual scene in Table~\ref{tab:qual_geo_full}.

\section{Details of scene parameters}

As mentioned in the main paper, we set $\mathbf{n}_f$ as $\langle0,0,1\rangle$ and fix it during training. In practice, capturing RGB images using mobile platforms with a gravity sensor (e.g. Apple ARKit and Android ARCore) can easily ensure that the estimated camera poses are aligned in z-axis with real world coordinate. However, it is difficult to ensure the normal directions in wall regions to be aligned with x/y-axis, that is the reason why we optimize $\mathbf{n}_w$ together with network parameters during training.

\begin{figure}[h]
\centering
\includegraphics[width=\linewidth]{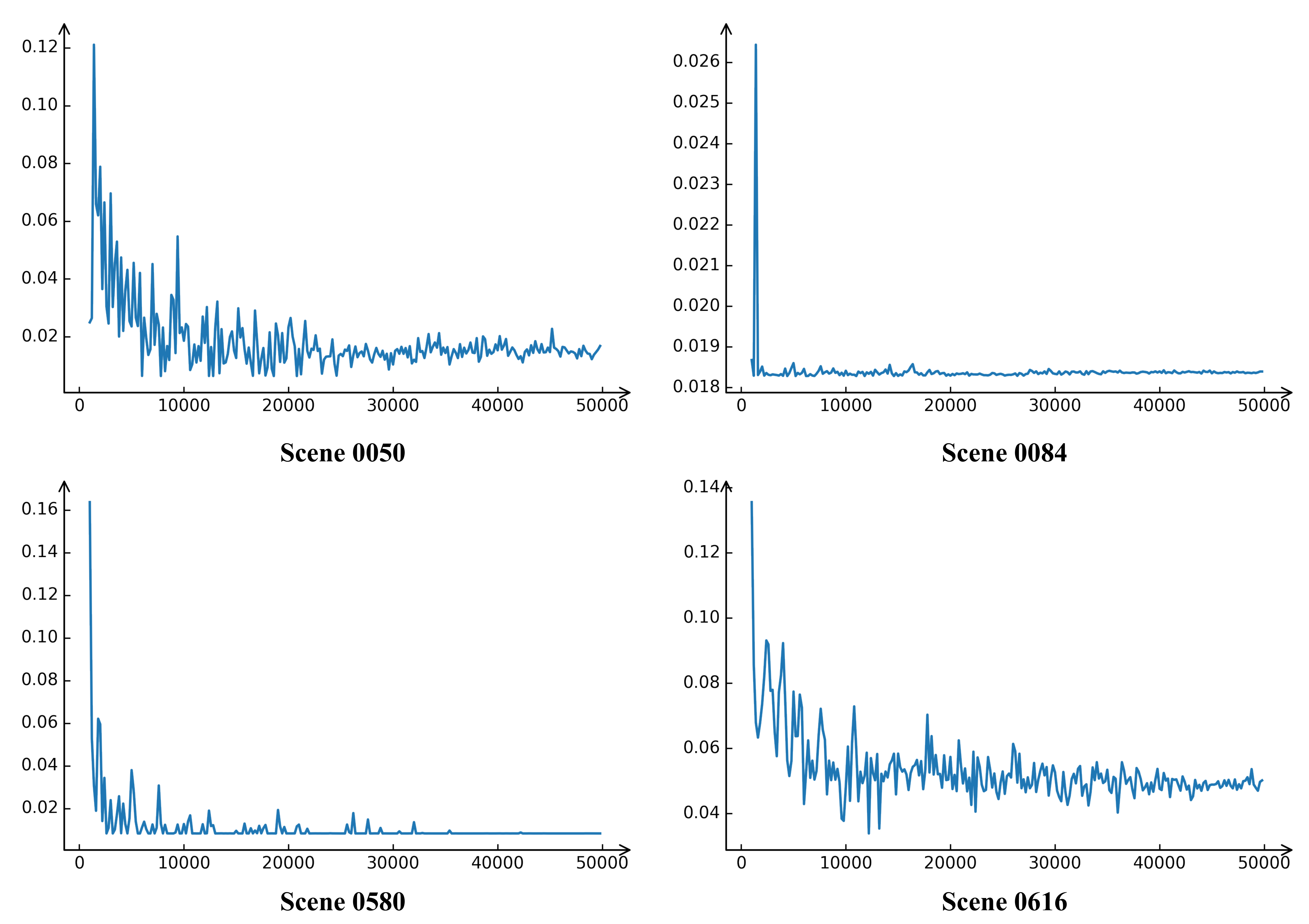}
\caption{\textbf{Convergence of $\mathbf{n}_w$.} x-axis represents training iteration number, y-axis represents the value of cost function in Equation~\eqref{eq:normal}.}
\label{fig:nw}
\end{figure}

In our experiments, we found that $\mathbf{n}_w$ can converge well. Since GT mesh with semantic label can be obtained on ScanNet dataset, we cluster the normal directions of wall regions using Mean shift algorithm and get 4 clustering centers $\{\mathbf{n}_i\}, i \in \{1,2,3,4\}$ for each scene. To evaluate how well $\mathbf{n}_w$ converges, we define cost function as:

\vspace{-0.5cm}

\begin{equation}
    L_\text{normal}(\mathbf{n}_w)=\frac{1}{4}\sum_{i=1}^{4}\min_{j\in\{-1,0,1\}}\left|j- \mathbf{n}_w \cdot \mathbf{n}_i\right|.
\label{eq:normal}
\end{equation}
The curves of cost function $L_\text{normal}(\mathbf{n}_w)$ are ploted in Figure~\ref{fig:nw}.

\vspace{-0.1cm}

\begin{table*}[t]
\vspace{2cm}
\centering
\scalebox{1.0}{
\begin{tabular}{|c|l|ccccc|ccccc|}
\hline

\parbox[t]{2mm}{\multirow{20}{*}{\rotatebox[origin=c]{90}{ScanNet}}}

& \multirow{2}{*}{Method} & \multicolumn{5}{c|}{scene 0050} & \multicolumn{5}{c|}{scene 0084} \\
\hhline{|~|~|----------|}
& & \multicolumn{1}{c}{Acc$\downarrow$} & \multicolumn{1}{c}{Comp$\downarrow$} & \multicolumn{1}{c}{Prec$\uparrow$} & \multicolumn{1}{c}{Recall$\uparrow$} & \multicolumn{1}{c|}{\cellcolor{gray!25}\textbf{F-score}$\uparrow$} & \multicolumn{1}{c}{Acc$\downarrow$} & \multicolumn{1}{c}{Comp$\downarrow$} & \multicolumn{1}{c}{Prec$\uparrow$} & \multicolumn{1}{c}{Recall$\uparrow$} & \multicolumn{1}{c|}{\cellcolor{gray!25}\textbf{F-score}$\uparrow$} \\

\hhline{|~|-----------|}

& COLMAP   & 0.059 & 0.174 & 0.659 & 0.491 & \cellcolor{gray!25} 0.563  & 0.042 & 0.134 & 0.736 & 0.552 & \cellcolor{gray!25} 0.631  \\
& COLMAP*  & 0.511 & 0.070 & 0.222 & 0.587 & \cellcolor{gray!25} 0.322  & 0.239 & 0.052 & 0.430 & 0.702 & \cellcolor{gray!25} 0.533  \\
& ACMP     & 0.123 & 0.097 & 0.560 & 0.594 & \cellcolor{gray!25} 0.577  & 0.122 & 0.060 & 0.567 & 0.652 & \cellcolor{gray!25} 0.607  \\
& NeRF     & 0.855 & 0.089 & 0.146 & 0.476 & \cellcolor{gray!25} 0.224  & 0.908 & 0.249 & 0.107 & 0.197 & \cellcolor{gray!25} 0.139  \\
& UNISURF  & 0.485 & 0.102 & 0.258 & 0.432 & \cellcolor{gray!25} 0.323  & 0.638 & 0.247 & 0.189 & 0.326 & \cellcolor{gray!25} 0.239  \\
& NeuS     & 0.130 & 0.115 & 0.441 & 0.406 & \cellcolor{gray!25} 0.423  & 0.255 & 0.360 & 0.128 & 0.084 & \cellcolor{gray!25} 0.101  \\
& VolSDF   & 0.092 & 0.079 & 0.512 & 0.544 & \cellcolor{gray!25} 0.527  & 0.551 & 0.162 & 0.127 & 0.232 & \cellcolor{gray!25} 0.164  \\ \hhline{|~|-----------|}
& Ours     & 0.058 & 0.059 & 0.707 & 0.642 & \cellcolor{gray!25} 0.673  & 0.055 & 0.053 & 0.639 & 0.621 & \cellcolor{gray!25} 0.630  \\
\hhline{|~|-----------|}

& \multirow{2}{*}{Method} & \multicolumn{5}{c|}{scene 0580} & \multicolumn{5}{c|}{scene 0616} \\
\hhline{|~|~|----------|}
& & \multicolumn{1}{c}{Acc$\downarrow$} & \multicolumn{1}{c}{Comp$\downarrow$} & \multicolumn{1}{c}{Prec$\uparrow$} & \multicolumn{1}{c}{Recall$\uparrow$} & \multicolumn{1}{c|}{\cellcolor{gray!25}\textbf{F-score}$\uparrow$} & \multicolumn{1}{c}{Acc$\downarrow$} & \multicolumn{1}{c}{Comp$\downarrow$} & \multicolumn{1}{c}{Prec$\uparrow$} & \multicolumn{1}{c}{Recall$\uparrow$} & \multicolumn{1}{c|}{\cellcolor{gray!25}\textbf{F-score}$\uparrow$} \\
\hhline{|~|-----------|}

& COLMAP   & 0.034 & 0.176 & 0.809 & 0.465 & \cellcolor{gray!25} 0.590  & 0.054 & 0.457 & 0.638 & 0.256 & \cellcolor{gray!25} 0.365  \\
& COLMAP*  & 0.479 & 0.054 & 0.223 & 0.661 & \cellcolor{gray!25} 0.333  & 0.354 & 0.147 & 0.211 & 0.429 & \cellcolor{gray!25} 0.283  \\
& ACMP     & 0.088 & 0.070 & 0.557 & 0.594 & \cellcolor{gray!25} 0.575  & 0.138 & 0.095 & 0.438 & 0.486 & \cellcolor{gray!25} 0.460  \\
& NeRF     & 0.457 & 0.138 & 0.099 & 0.252 & \cellcolor{gray!25} 0.142  & 0.718 & 0.233 & 0.172 & 0.237 & \cellcolor{gray!25} 0.199  \\
& UNISURF  & 0.376 & 0.116 & 0.218 & 0.399 & \cellcolor{gray!25} 0.282  & 0.716 & 0.193 & 0.183 & 0.293 & \cellcolor{gray!25} 0.225  \\
& NeuS     & 0.161 & 0.215 & 0.413 & 0.327 & \cellcolor{gray!25} 0.365  & 0.171 & 0.142 & 0.269 & 0.284 & \cellcolor{gray!25} 0.276  \\
& VolSDF   & 0.091 & 0.088 & 0.529 & 0.540 & \cellcolor{gray!25} 0.534  & 0.922 & 0.150 & 0.115 & 0.259 & \cellcolor{gray!25} 0.160  \\ \hhline{|~|-----------|}
& Ours     & 0.104 & 0.062 & 0.616 & 0.650 & \cellcolor{gray!25} 0.632  & 0.072 & 0.098 & 0.521 & 0.431 & \cellcolor{gray!25} 0.472  \\
\hline
\hline

\parbox[t]{2mm}{\multirow{20}{*}{\rotatebox[origin=c]{90}{7-Scenes}}}

& \multirow{2}{*}{Method} & \multicolumn{5}{c|}{Heads} & \multicolumn{5}{c|}{Office} \\
\hhline{|~|~|----------|}
& & \multicolumn{1}{c}{Acc$\downarrow$} & \multicolumn{1}{c}{Comp$\downarrow$} & \multicolumn{1}{c}{Prec$\uparrow$} & \multicolumn{1}{c}{Recall$\uparrow$} & \multicolumn{1}{c|}{\cellcolor{gray!25}\textbf{F-score}$\uparrow$} & \multicolumn{1}{c}{Acc$\downarrow$} & \multicolumn{1}{c}{Comp$\downarrow$} & \multicolumn{1}{c}{Prec$\uparrow$} & \multicolumn{1}{c}{Recall$\uparrow$} & \multicolumn{1}{c|}{\cellcolor{gray!25}\textbf{F-score}$\uparrow$} \\
\hhline{|~|-----------|}

& COLMAP   & 0.036 & 0.284 & 0.815 & 0.206 & \cellcolor{gray!25} 0.329  & 0.080 & 0.247 & 0.408 & 0.195 & \cellcolor{gray!25} 0.264  \\
& COLMAP*  & 0.921 & 0.299 & 0.110 & 0.137 & \cellcolor{gray!25} 0.122  & 0.769 & 0.143 & 0.125 & 0.251 & \cellcolor{gray!25} 0.167  \\
& ACMP     & 0.099 & 0.220 & 0.427 & 0.259 & \cellcolor{gray!25} 0.322  & 0.261 & 0.180 & 0.259 & 0.203 & \cellcolor{gray!25} 0.228  \\
& NeRF     & 0.144 & 0.323 & 0.294 & 0.047 & \cellcolor{gray!25} 0.081  & 0.669 & 0.341 & 0.185 & 0.079 & \cellcolor{gray!25} 0.111  \\
& UNISURF  & 0.129 & 0.117 & 0.354 & 0.419 & \cellcolor{gray!25} 0.384  & 0.534 & 0.145 & 0.127 & 0.248 & \cellcolor{gray!25} 0.168  \\
& NeuS     & 0.091 & 0.204 & 0.520 & 0.328 & \cellcolor{gray!25} 0.402  & 0.174 & 0.242 & 0.203 & 0.135 & \cellcolor{gray!25} 0.162  \\
& VolSDF   & 0.136 & 0.097 & 0.396 & 0.417 & \cellcolor{gray!25} 0.406  & 0.412 & 0.147 & 0.139 & 0.195 & \cellcolor{gray!25} 0.162  \\ \hhline{|~|-----------|}
& Ours     & 0.120 & 0.086 & 0.407 & 0.428 & \cellcolor{gray!25} 0.417  & 0.116 & 0.165 & 0.254 & 0.172 & \cellcolor{gray!25} 0.205  \\
\hhline{|~|-----------|}

& \multirow{2}{*}{Method} & \multicolumn{5}{c|}{Chess} & \multicolumn{5}{c|}{Fire} \\
\hhline{|~|~|----------|}
& & \multicolumn{1}{c}{Acc$\downarrow$} & \multicolumn{1}{c}{Comp$\downarrow$} & \multicolumn{1}{c}{Prec$\uparrow$} & \multicolumn{1}{c}{Recall$\uparrow$} & \multicolumn{1}{c|}{\cellcolor{gray!25}\textbf{F-score}$\uparrow$} & \multicolumn{1}{c}{Acc$\downarrow$} & \multicolumn{1}{c}{Comp$\downarrow$} & \multicolumn{1}{c}{Prec$\uparrow$} & \multicolumn{1}{c}{Recall$\uparrow$} & \multicolumn{1}{c|}{\cellcolor{gray!25}\textbf{F-score}$\uparrow$} \\
\hhline{|~|-----------|}

& COLMAP   & 0.112 & 0.772 & 0.271 & 0.119 & \cellcolor{gray!25} 0.165  & 0.047 & 0.365 & 0.652 & 0.287 & \cellcolor{gray!25} 0.399  \\
& COLMAP*  & 0.373 & 0.197 & 0.117 & 0.180 & \cellcolor{gray!25} 0.142  & 0.616 & 0.220 & 0.114 & 0.293 & \cellcolor{gray!25} 0.164  \\
& ACMP     & 0.747 & 0.194 & 0.158 & 0.211 & \cellcolor{gray!25} 0.181  & 0.064 & 0.181 & 0.555 & 0.402 & \cellcolor{gray!25} 0.466  \\
& NeRF     & 0.502 & 0.405 & 0.118 & 0.061 & \cellcolor{gray!25} 0.081  & 0.979 & 0.216 & 0.038 & 0.153 & \cellcolor{gray!25} 0.061  \\
& UNISURF  & 0.285 & 0.160 & 0.208 & 0.281 & \cellcolor{gray!25} 0.239  & 0.682 & 0.121 & 0.092 & 0.256 & \cellcolor{gray!25} 0.136  \\
& NeuS     & 0.206 & 0.404 & 0.199 & 0.136 & \cellcolor{gray!25} 0.162  & 0.134 & 0.139 & 0.330 & 0.316 & \cellcolor{gray!25} 0.323  \\
& VolSDF   & 0.364 & 0.180 & 0.148 & 0.232 & \cellcolor{gray!25} 0.181  & 0.229 & 0.135 & 0.197 & 0.296 & \cellcolor{gray!25} 0.236  \\ \hhline{|~|-----------|}
& Ours     & 0.129 & 0.214 & 0.289 & 0.244 & \cellcolor{gray!25} 0.265  & 0.083 & 0.066 & 0.455 & 0.460 & \cellcolor{gray!25} 0.458  \\
\hline

\end{tabular}
}
\caption{\textbf{3D reconstruction metrics of individual scenes on ScanNet and 7-Scenes.}}
\vspace{2cm}
\label{tab:qual_geo_full}
\end{table*}

\section{Novel View synthesis results}

\begin{figure}[h]
\centering
\includegraphics[width=\linewidth]{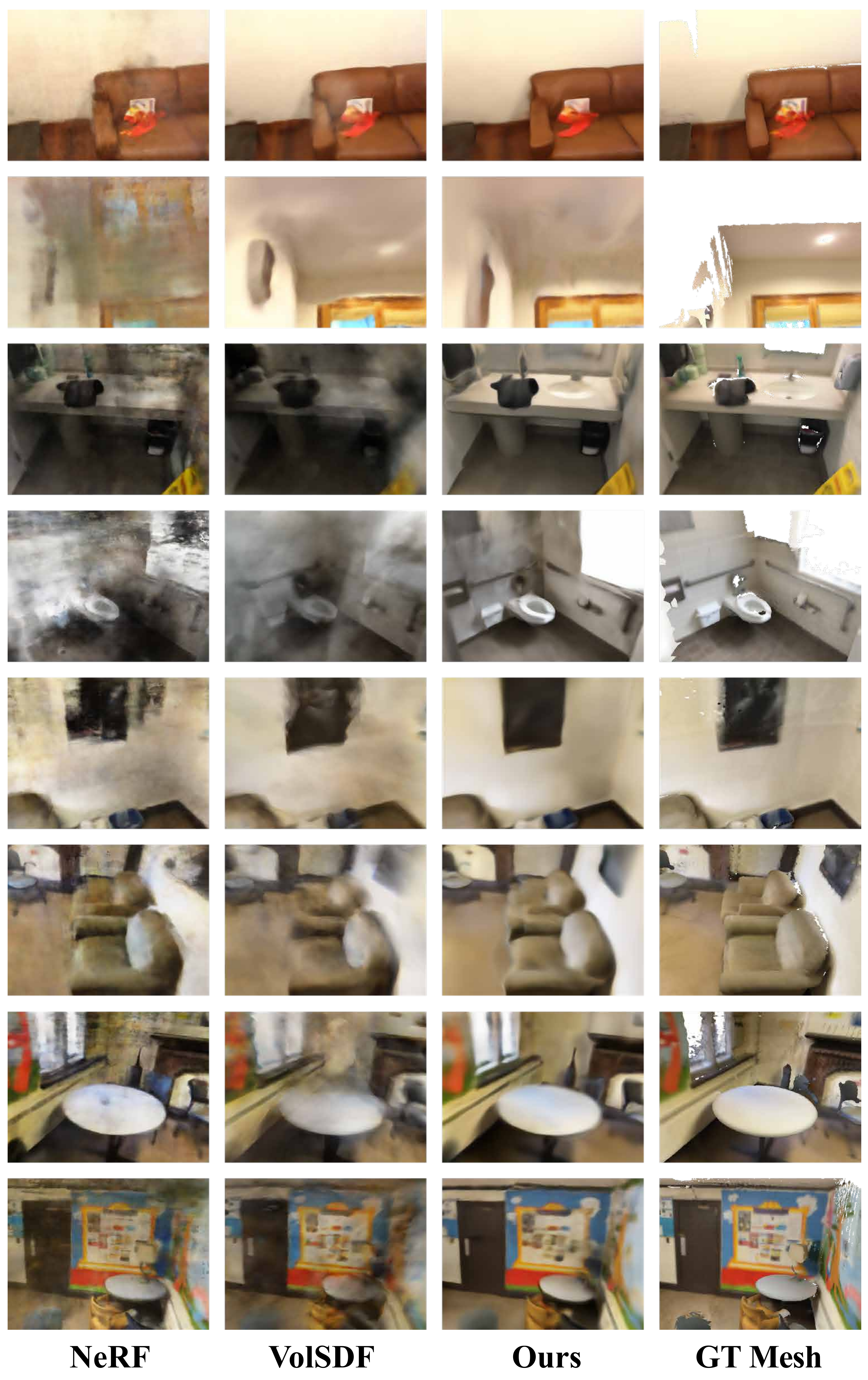}
\caption{\textbf{Novel view synthesis results on ScanNet.}  Due to the lack of ground truth images in novel views, we render ground truth mesh in these views for reference.}
\label{fig:nv}
\end{figure}

To evaluate the performance of our method on novel view synthesis, we manually set novel camera poses by adding 0.2 to training camera poses along the z-axis. Figure~\ref{fig:nv} presents the qualitative comparisons on ScanNet. Please refer to the supplementary video for more results.

\section{Performance on non-Manhattan scenes.}
Our method is not suitable to non-Manhattan scenes due to there may exist sloped ground and walls maybe not vertical to each other, making our geometric constraints cannot work well.

\section{Zoom-in visualization of non-planar regions.}
\begin{figure}[h]
    \centering
    \includegraphics[width=\linewidth]{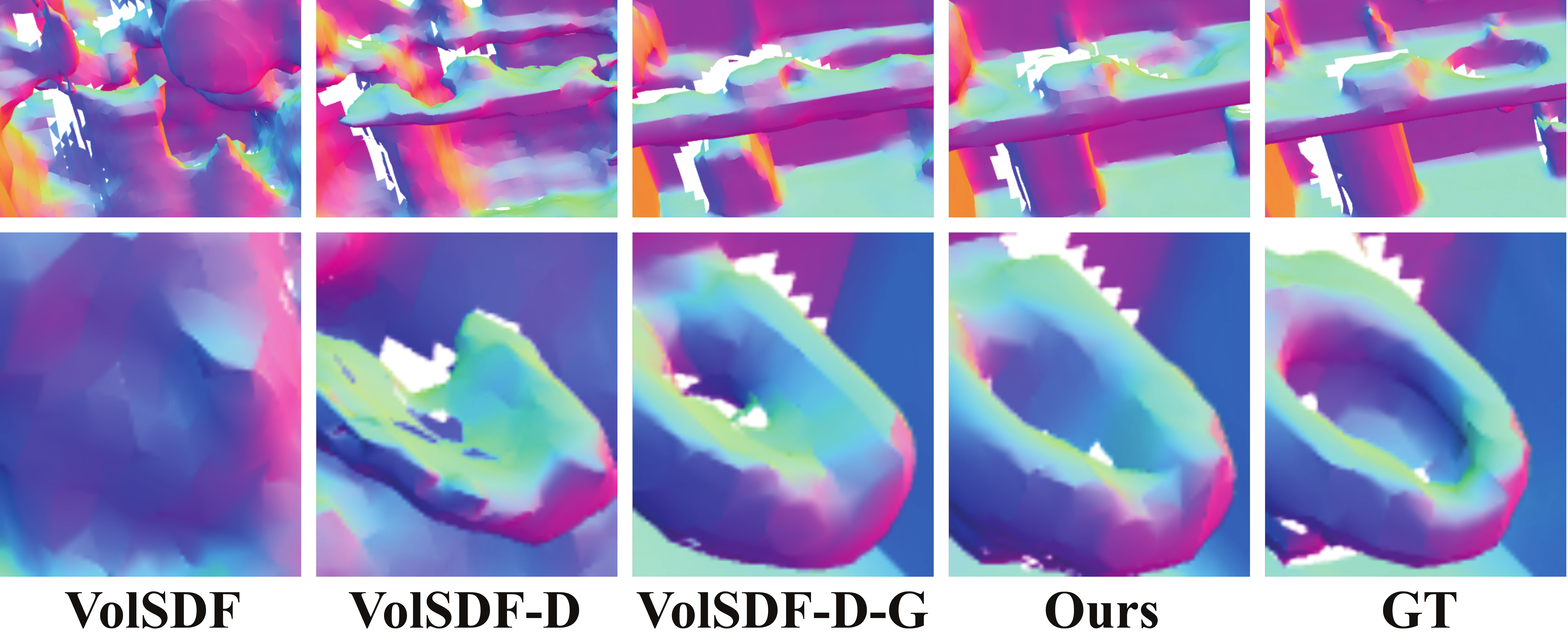}
    \caption{\textbf{Zoom-in visualization of non-planar regions.}}
    \label{fig:non_planar}
\end{figure}

We provide zoom-in visualization of geometric details of non-planar regions in Fig.~\ref{fig:non_planar}.

\section{Effect of inaccurate segmentation.}
\begin{figure}[h]
    \centering
    \includegraphics[width=\linewidth]{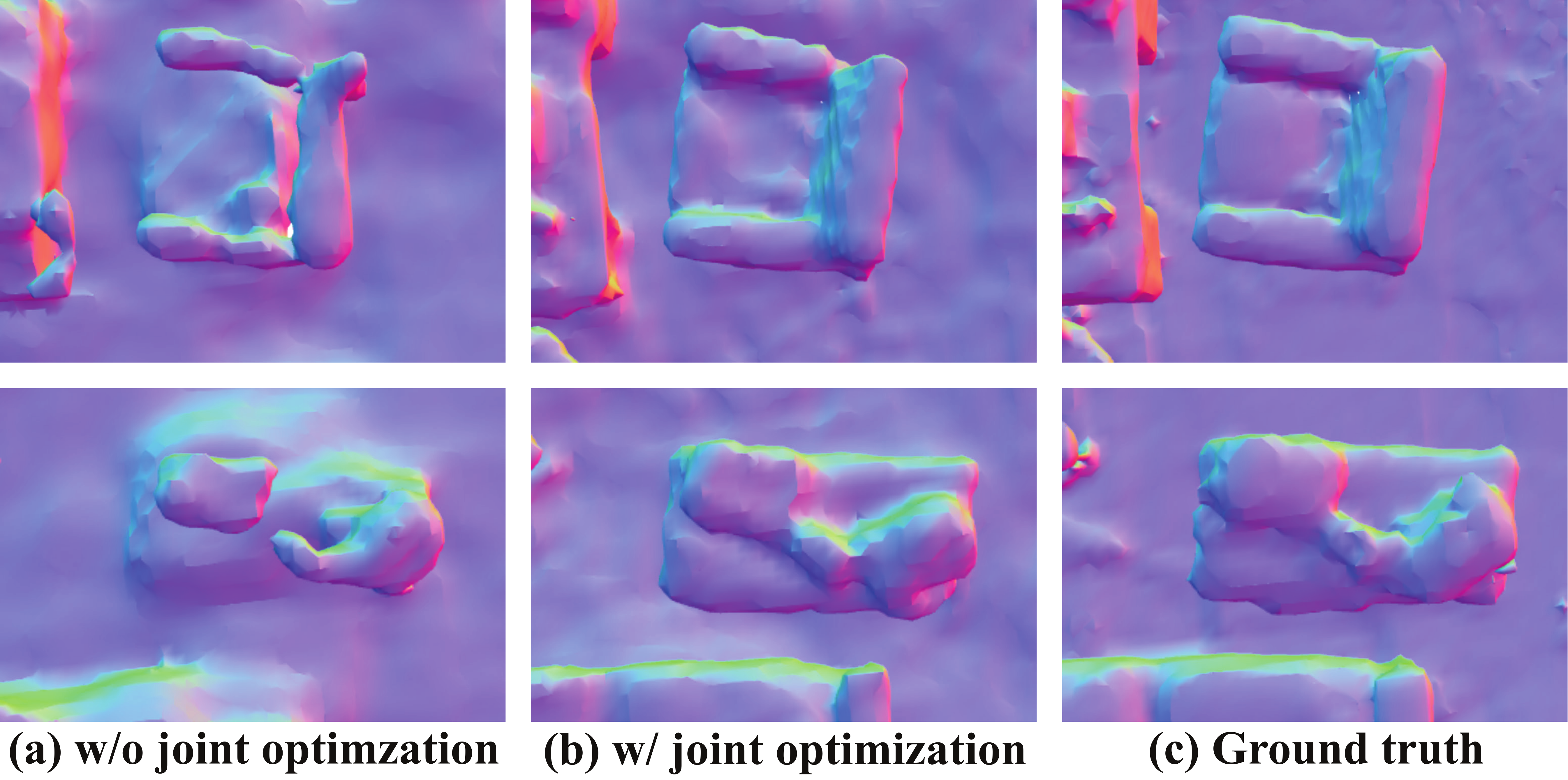}
    \caption{\textbf{Effect of inaccurate segmentation.}}
    \label{fig:inaccurate_seg}
\end{figure}

If non-planar regions are mis-classified as planar, the reconstruction may be misled by the wrong regularization as shown in Fig.~\ref{fig:inaccurate_seg} (a). Our joint optimization can correct it as shown in Fig.~\ref{fig:inaccurate_seg} (b).

\section{Why non-planar regions are improved?}
As the whole scene is represented by a single SDF network with the Eikonal regularization, the convergence and reconstruction quality of planar regions (dominant area) will also influence those in non-planar regions.

\section{Comparison with supervised method.}
We provide results of DeepVideoMVS in Tab.~\ref{tab:more_baseline}.

\begin{table}[h]\scriptsize
    \centering
    \begin{tabular}{lccccc}
    \hline
    & Acc$\downarrow$ & Comp$\downarrow$ & Prec$\uparrow$ & Recall$\uparrow$ & \cellcolor[gray]{0.902}F-score$\uparrow$ \\
    \hline
    DeepVideoMVS & 0.206 & \textbf{0.032} & 0.286 & \textbf{0.827} & \cellcolor[gray]{0.902}0.424  \\
    Ours         & \textbf{0.072} & 0.068 & \textbf{0.621} & 0.586 & \cellcolor[gray]{0.902}\textbf{0.602}  \\
    \hline
    \end{tabular}
    \caption{\textbf{Comparison with DeepVideoMVS on ScanNet.}}
    \label{tab:more_baseline}
\end{table}

\end{document}